\documentclass[11pt]{article}

\usepackage[final]{acl}

\usepackage{times}
\usepackage{latexsym}

\usepackage[T1]{fontenc}
\usepackage[utf8]{inputenc}

\usepackage{microtype}

\usepackage{inconsolata}

\usepackage{algorithm}
\usepackage{algorithmic}
\usepackage{amssymb}
\usepackage{amsmath}
\usepackage{array}
\usepackage{booktabs}
\usepackage{multirow}
\usepackage{multicol}
\usepackage{makecell}
\usepackage{graphicx}

\title{Annotation-Free Reinforcement Learning Query Rewriting \\
via Verifiable Search Reward}

\author{
    Sungguk Cha, DongWook Kim, Taeseung Hahn, Mintae Kim, Youngsub Han, Byoung-Ki Jeon \\
    LG Uplus, South Korea \\
    \texttt{\{sungguk, dongwook92, tshahn, iammt, yshan042, bkjeon\}@lguplus.co.kr}
}

\begin{document}
\maketitle
\begin{abstract}
Optimizing queries for Retrieval-Augmented Generation (RAG) systems poses a significant challenge, particularly across diverse modal indices.
We introduce RL-QR, a novel annotation-free reinforcement learning framework for query rewriting that eliminates the need for costly human-annotated data.
By leveraging verifiable search rewards derived from index-aligned synthetic queries, RL-QR overcomes human-annotation dependencies, extending its applicability to various modalities and index domains.
Experimental results demonstrate the framework's robustness, achieving substantial retrieval performance gains of up to 3.9$\times$ on lexical retrievers and 3.5$\times$ on semantic retrievers on the MTEB VIDORE V2 benchmark for unstructured visual documents, along with consistent 5\% to 10\% improvements on MS MARCO v2.1 and internal industrial datasets.
\end{abstract}
\section{Introduction}
Retrieval-Augmented Generation (RAG)~\cite{lewis2020retrieval} has proven to be a powerful and widely adopted approach across numerous domains, from natural language processing to multi-modal applications. 
Its ability to integrate external knowledge into generation tasks has made it a cornerstone of modern retrieval systems. 
Modern AI assistants~\cite{hurst2024gpt, comanici2025gemini} adopt RAG as core function for correcting factually, delivering out-domain knowledge and beyond. 

In practice, when serving RAG systems across various domains and index formats, adapting queries through rewriting proves to be more effective and cost-efficient than rebuilding retrievers.
For lexical retrievers, creating domain-specific dictionaries can enhance performance. 
However, this approach depends on manual annotation, which is not scalable and increases operational costs.
For semantic indices, retrievers can be fine-tuned with domain-specific data. 
Yet, this introduces the burden of maintaining domain-specific retrievers, generating training data, and conducting retraining. 
Moreover, updating retrievers typically requires re-indexing, which adds complexity to RAG system dependencies and further raises operational costs.
In contrast, query rewriters transform queries into the representation space of the retrievers, allowing compatibility across different retrievers and index types.
From a system maintenance and deployment perspective, developing a query rewriter is generally more cost-effective than enhancing retrievers or re-indexing. 
It also promotes modularity in RAG architecture by decoupling the query rewriting module from retriever components, avoiding the need for domain-specific retriever development.

Although query rewriting is central to RAG systems, generalized approaches remain largely unexplored due to their reliance on costly human annotation.
Recent studies have proposed implicit learning methods that reward the model when the final answer is correct, requiring annotated index-query-answer-verifier sets~\cite{jin2025search}.
Others use explicit learning, which rewards the model when relevant documents are retrieved, but this approach depends on expensive per-query annotations of both positive and negative document pairs~\cite{wang2025maferw}.
While these methods show promise within narrow domains, they face major limitations: they require extensive human effort and are largely restricted to curated, text-only data sources—making them unsuitable for real-world, unstructured document collections.

This study introduces an annotation-free \textbf{R}einforcement \textbf{L}earning framework for \textbf{Q}uery \textbf{R}ewriting with verifiable search reward (\textbf{RL-QR}).
The framework's core novelty lies in synthesizing queries in an index-aligned manner, which permits the proposed verifiable search reward to directly exploit the resultant search score for training. 
Crucially, this methodology replaces the need for positive query-corpus human-annotations, thus ensuring off-the-shelf operation and achieving general applicability across different index modalities and domains.

Our experiments demonstrate robust and significant improvements on text-modal and multi-modal unstructured documents with the RAG agent framwork.
Especially, upon the general visual document retrieval benchmark MTEB VIDORE V2~\cite{macé2025vidorebenchmarkv2raising} the conventional text-parsing based RAG system benefits upto $3.9\times$ retrieval recall.
For the other benchmarks, the text-modal retrieval benchmark MS MARCO v2.1~\cite{DBLP:journals/corr/NguyenRSGTMD16} and internal industrial unstructured document benchmark, RL-QR steadily achieves 5\% to 10\% performance gains. 
It supports the effectiveness and the adaptability over various index domains and modalities.

In summary, our contributions are

\begin{itemize}
    \item \textbf{Annotation-Free RL Framework with Verifiable Rewards:} We introduce RL-QR, a novel reinforcement learning framework that eliminates the dependency on costly human-annotated data for query rewriting.
    By leveraging index-aligned synthetic queries to generate verifiable search rewards, our approach explicitly optimizes the rewriter using the resultant search scores.

    \item \textbf{Universal Adaptability and Modular System Integration:} We propose a retriever- and index-agnostic solution that ensures general applicability across diverse modalities, including text-parsed and unstructured visual documents. 
    This modular design decouples the query rewriting module from the retriever, significantly reducing system maintenance overhead by removing the need for domain-specific retriever development or expensive re-indexing processes.

    \item \textbf{Robust Empirical Effectiveness:} Extensive experiments demonstrate that RL-QR achieves substantial performance gains, proving its efficacy in "unlearning" chat-oriented behaviors to prioritize retrieval intent, including up to 3.9x improvement on lexical retrievers and 3.5x on semantic retrievers within the MTEB VIDORE V2 benchmark.
    Additionally, the framework consistently delivers 5\% to 10\% recall enhancements on the MS MARCO v2.1 benchmark and internal industrial datasets, proving its efficacy in "unlearning" chat-oriented behaviors to prioritize retrieval intent.
\end{itemize}

\section{Related Works}

Our work focuses on enhancing the query rewriter for RAG systems, with an emphasis on handling multi-modal (unstructured imaged documents) and text-modal (text-parsed documents) indices with real-world unstructured data. 
In this section, we provide an overview of the research background, covering the evolution of RAG, the integration of various modalities in RAG systems, and the role of query rewriting.

\subsection{Retrieval-Augmented Generation (RAG)}

RAG is a hybrid approach that integrates retrieval-based and generation-based techniques to improve the performance of language models on knowledge-intensive tasks. 
By leveraging external knowledge sources, RAG enables models to produce more accurate and contextually relevant responses. 
The paradigm has gained significant attention due to its ability to combine the strengths of retrieving pertinent documents and generating coherent text.
In the real-world, RAG systems are widely adopted with online web search (\textit{e.g.,} OpenAI~\cite{hurst2024gpt}, and Gemini~\cite{comanici2025gemini}) and industrial domains with credential documents. 

Early research on RAG established its effectiveness across various natural language processing tasks~\cite{lewis2020retrieval}. 
Subsequent studies have proposed advancements, such as improved retrieval mechanisms using dense retrieval methods~\cite{karpukhin2020dense} and the integration of structured knowledge bases like databases or graphs~\cite{edge2024local}. 
RAG has also shown promise in multi-task and few-shot learning scenarios~\cite{izacard2023atlas}, where the retrieval component compensates for limited training data by accessing external information. 
However, challenges remain, particularly in optimizing the retrieval process, which depends heavily on the quality of the input query—an issue that motivates the exploration of query rewriting~\cite{ma2023query}.

\subsection{Modalities in RAG}

While RAG was initially designed for text-based applications, recent applications have extended their scope to the real-world unstructured documents including slide decks, web pages, blogs, papers and so on supported by document parsing approaches. 
This expansion is critical for tasks where knowledge sources span multiple formats, requiring systems to integrate and reason over heterogeneous inputs.

Multi-modal RAG systems have been explored for image-as-embedding~\cite{faysse2024colpali} or parsing-documents-to-text~\cite{feng2025dolphin}.
Image-as-embedding approaches~\cite{faysse2024colpali} embeds imaged document into document embedding as document semantic embedding~\cite{qwen3embedding}.
Parsing-documents-to-text approaches~\cite{wei2024general, feng2025dolphin} converts documents into plain text, enabling the present text retrievers~\cite{robertson2009probabilistic, qwen3embedding}.
For text-modal data, such as parsed text from structured documents, the challenge lies in effectively retrieving and utilizing information from long-form or hierarchically organized content~\cite{larson2024graphrag}.

\subsection{Query Rewriting for RAG}

Query rewriting is a pivotal component in RAG systems, as the effectiveness of the retrieval step hinges on how well the query is formulated. 
A poorly designed query can lead to irrelevant or low-quality retrieved documents, undermining the generation process. 
Conversely, an optimized query enhances the relevance of retrieved information, directly improving the overall system performance.

Traditional query rewriting techniques, such as query expansion and reformulation, have roots in information retrieval and rely on heuristics or statistical methods to refine queries~\cite{zhu2016query}. 
In the context of RAG, however, query rewriting must align with the needs of the retriever~\cite{ma2023query}. 
Recent efforts have introduced learning-based approaches, including neural network models and reinforcement learning, to dynamically adapt queries based on system feedback~\cite{wang2025maferw, chan2024rq, li2024dmqr, ma2023query, jin2025search}. 
Despite these advances, existing query rewriting techniques often require extensive annotated data or are constrained to specific domains~\cite{liu2021conversational}. 
Our work addresses these gaps by developing a generalized reinforcement learning framework for query rewriting, tailored to enhance retrieval across diverse indices without relying on large-scale human annotations. 
\section{Method}
In this section, we describe our proposed framework, annotation-free reinforcement learning query rewriting via verifiable search reward (RL-QR).
Illustrated in the Alg.~\ref{alg:rlqr}, the learning process consists of two-steps:
(1) index-algined query synthesis where the answer for the query necessarily requires the source index resulting the index matching queries,
(2) and reinforcement learning the query rewriter based on the verifiable search reward.

\begin{algorithm}[ht]
    \caption{Annotation-free RL-QR}
\label{alg:rlqr}
\begin{algorithmic}[1]
\REQUIRE Retriever $R$, Search Index $\mathcal{I}$, Query synthesis assistant $Assistant$
\ENSURE Optimized Policy $\pi_\theta$
\STATE Initialize $\pi_\theta$
\FOR{index $I$ in $\mathcal{I}$}
    \STATE $q \leftarrow Assistant(I)$
    \STATE $rewards \leftarrow EmptyList$
    \FOR{for $i$ in $N_{rollout}$}
        \STATE $q'_i \leftarrow \pi_\theta(q)$
        \STATE $SearchedIndices_i \leftarrow R(q'_i)$
        \STATE $r_i \leftarrow NDCG(I, SearchedIndices_i)$
        \STATE $rewards$.append($r_i$)
    \ENDFOR
    \STATE $A \leftarrow GroupComputation(rewards)$
    \STATE Update policy $\pi_\theta$ using advantages $A$
\ENDFOR
\RETURN $\pi_\theta$
\end{algorithmic}
\end{algorithm}

\subsection{Index-algined Query Synthesis}
Query synthesis have already been explored widely~\cite{xu2024magpiealignmentdatasynthesis, cha2024visuallydehallucinativeinstructiongeneration}, and the modern language models are capable enough to conduct the task. 
In this work, we generate queries for training in index-aligned manner where the answer for the query necessarily requires the corpus (see the prompt provided in Table\ref{tab:prompt}).

\begin{table}[!h]
\begin{tabular}{p{0.9\columnwidth}}
\toprule
\textbf{\# Generating Document-Requiring Question and Answer} \\
\midrule
Read the document carefully, then perform the following three steps:
\begin{enumerate}
    \item Think of a **scenario** that necessitates the information contained within the document.
    \item Create a **question** that logically fits the identified scenario.
    \item Provide an **answer** that accurately matches the created question based on the document's content.
\end{enumerate}
% \addlinespace
\textit{Note: If the document's information is insufficient to identify a situation requiring the document, output blank spaces.}
\\
% \midrule
% \textbf{Final Response Format:} \\
% <scenario>...</scenario> \\
% <question>...</question> \\
% <answer>...</answer> \\
\bottomrule
\end{tabular}
\caption{Prompt template for index-algined query synthesis. Appending the resulting scenario and question is viable to make longer queries. Training the rewriter does not utilize the generated answers, but those are useful for post-training the answer model and evaluating end-to-end RAG.}
\label{tab:prompt}
\end{table}

Specifically, for the conventional text-parsing retrieval system, given the raw data $DB_{raw}$, the document parsers $P$ and the search engine $E$, the source index $\mathcal{I}_{text}$ becomes
\begin{equation}
    \mathcal{I}_{text} \leftarrow \bigcup_{d \in DB_{raw}}{E.index(P.parse(d))}
\end{equation}
which contains both the search engine index and the parsed corpus.
On the other hand, the multimodal retrieval system does not adopt the parser.
Consequently, the multimodal index $\mathcal{I}_{multimodal}$ becomes
\begin{equation}
    \mathcal{I}_{multimodal} \leftarrow \bigcup_{d \in DB_{raw}}{E.index(d)}
\end{equation}
where $E$ utilizes multimodal embedding model (e.g., \cite{faysse2024colpali}) for raw document indexing.
Along with the query synthesis assistant (e.g., \cite{bai2025qwen3vltechnicalreport}) and the instruction (Table~\ref{tab:prompt}), it  generates index-aligned queries per index as illustrated in Alg.~\ref{alg:rlqr}, which can be done online and offline train.

\subsection{Reinforcement Learning Query Rewriter}
It is important to individualize query rewriter with respect to the indices, because each retriever has distinct characteristics.
For example, lexical retrievers such as BM25~\cite{robertson2009probabilistic} count on the number of the words, in which simply repeating important word can augment the performance. 
Whereas, (multi-modal) semantic retrievers that embed (text-parsed-)documents into embedding~\cite{faysse2024colpali, qwen3embedding} work better if the query-document resembles their trained data, which is hard to manage. 
The reinforcement learning aims to align user query into the index representation space by the query rewriter per retriever $R$ and its index $\mathcal{I}$.
In other words, for $N$ online RAG systems consisting of the data source $DB_i$ and the retriever $R_i$ for $i \in N$, we suggest to have $N$ rewriters respectively, rather than a single universal rewriter. 

The precedent RL approaches~\cite{ma2023query, jin2025search, nguyen2025rl} implicitly train the rewriter by optimizing
\begin{equation}
    \max_{\pi_{\theta}, \pi_{LLM}} \mathbb{E}_{x \sim D, y \sim \pi_{\theta}(\cdot | x; R), z \sim \pi_{LLM}(\cdot | x;R(y))} \left[ r_{\phi}(x, z) \right]
    \label{eq:rl_precedent}
\end{equation}
where $x$ refers to the sample from the training data $D$, $y$ denotes the rewritten query by the rewriter, and $z$ represent the final response.
$\pi_{\theta}$ and $\pi_{LLM}$ are the target rewriter and the final-responding language model.
$r_{\phi}$ is the reward function but inherently requires human-annotation in query-response or query-retrieval levels.

In contrast, ours optimizes the rewriter explicitly, which down-scales the objective and boosts the training process.
Some~\cite{wang2025maferw} tried explicit rewarding with massive document-wise positive and negative pair annotation, which limits in scaling covering domain and indices. 
On the other hand, leveraging the synthesized index-aligned queries, we formulate the RL objective function as follows:
\begin{equation}
    \max_{\pi_{\theta}} \mathbb{E}_{x \sim D, y \sim \pi_{\theta}(\cdot | x; R)} \left[ r_{\phi}(x, y) \right]
    \label{eq:rl}
\end{equation}
We adopt two function rewards, one for the query rewriting reward and the other for the formatting and redundant penalty.
The verifiable search reward $r_{\text{retrieval}}$ uses NDCG~\cite{jarvelin2002cumulated} score directly that measures if the target document is retrieved considering ranks.
\begin{equation}
    r_{\text{retrieval}}(x, y) = NDCG(\text{index}_x, R(y))
\end{equation}
The penalty $r_{penalty}$ targets to match the format, placing the rewritten query inside \verb|<answer>...</answer>|, and reduce redundant generations outside the format. 
It is similar to and replacable with the well known length penalties.
Further, we normalized the penalty $r_{\text{penalty}}$ group-wise by ranging \verb|[0.5, 1]| for the non-zero values.
% \begin{equation}
%     \text{penalty(y)} = 
%     \begin{cases}
%         |y| - |\text{formatted query}|, & \text{if format matched} \\
%         \infty, & \text{otherwise}
%     \end{cases}
% \end{equation}
% \begin{equation}
% r_{\text{penalty}}(y) =
% \begin{cases}
% 0, & \text{if } \texttt{penalty}(y) = 0 \\
% \texttt{redundancy}(y), & \text{otherwise}
% \end{cases}
% \end{equation}
The reward function becomes
\begin{equation}
    r_{\phi}(x, y) = \lambda_1 r_{\text{retrieval}(x, y)} + \lambda_2 r_{\text{penalty}(y)}
\end{equation} \label{eq:lambda}
where the lambdas are the hyper-parameters.
More specifically, for each sample $x$, we optimize the rewriter by maximizing the following object:
\begin{align}
\mathcal{J}_{\text{GRPO}}(\theta) 
&= \mathbb{E} \left[ x \sim D, \{o_i\}_{i=1}^{G} \sim \pi_{\theta_{\text{old}}}(O|q) \right] \nonumber \\
&\Bigg[
\frac{1}{G} \sum_{i=1}^{G} \frac{1}{|o_i|} \sum_{t=1}^{|o_i|}
\Bigg\{
\min \Bigg(
r_{i,t}(\theta) \hat{A}_{i,t}, \nonumber \\
&
\text{clip} \Bigg(
r_{i,t}(\theta),
1 - \epsilon, 1 + \epsilon
\Bigg) \hat{A}_{i,t}
\Bigg) \Bigg\} \Bigg]
\end{align} \label{eq:grpo}

where $r_{i,t}(\theta) = \frac{\pi_{\theta}(o_{i,t} \mid x, o_{i,<t})} {\pi_{\theta_{\text{old}}}(o_{i,t} \mid x, o_{i,<t})}$ be the probability ratio, $\epsilon$ and $\beta$ are hyper-parameters, and $\hat{A}_{i, t}$ is the advantage based on the relative rewards of the outputs inside each group.
\section{Experiment}

\subsection{RAG System Implementation}
In this experiment, we adopt three in-house RAG frameworks named \textit{Semantic}, \textit{Lexical} and \textit{Multimodal}.
\textit{Semantic} and \textit{Lexical} are the conventional RAG framework containing (1) the document parser that ingests unstructured documents and chunks into corpus (2) the embedding models and (3) the search engine. 
\textit{Multimodal} is the modern RAG framework without document parsers.

\textit{Semantic} utlizes an embedding model that converts the user query and the corpus into vector embeddings and searches by its search engine indexed with the corpus vector embeddings. 
This approach allows the system to understand the user intent and retrieve relevant information even if the phrasing is different and there is no direct keyword overlap.

\textit{Lexical} is a traditional information retrieval system built upon BM25 scoring algorithm which conducts retireval by matching the exact token between the user query and the corpus.
It excels at precision when the query contains specific terms, acronyms, or identifiers that are also present in the source documents, as its ranking is based on term frequency (TF) and inverse document frequency (IDF).

\textit{Multimodal} is conceptually identical to which of \textit{Semantic}, but has a difference in the raw document compatibility. 
% It can embed documents in both forms text-parsed or image-level.
As it does not have external parser which used to have information leaks, it shows stronger retrieval performance when it comes to unstructured documents (e.g., slide decks, pdf). 

\subsection{Experiment Data}
Experimental setup has two major categories: (1) text-only retrieval task and (2) unstructured document retrieval task.
For the text-only retrieval task, we use MS MARCO v2.1 testset 1\% which has 10,078 corpus and 1,011 queries.
The unstructured visual document retrieval task consists of MTEB VIDORE V2 benchmark with 4,544 visual documents and 327 queries; and in-house real-world industrial data composed of 2,791 documents and 4,398 queries. 

The virtue of retrieval task is to maximize recall, which NDCG represents it well with ordinal scoring.
Therefore, we deploy NDCG@3 for the target evaluation metric and the reward score. 

\subsection{Index-Aligned Query Synthesis}
We generated queries per index using Qwen3-VL-235B-A22B~\cite{bai2025qwen3vltechnicalreport} and assumed the generated query has single retrieval target, the source corpus.
In other words, the reward becomes if the query retrives the source corpus from the retriever.
% Specifically we generated 10,079 queries for MARCO, 1,538 queries for VidoreV2/ESGReportsRetrieval, 452 queries for VidoreV2/EconomicsReportsRetireval, 1,538 queries for VidoreV2/ESGReportsHLRetrieval, 1,016 queries for VidoreV2/BioMedicalLecturesRetrieval and 2,791 queries for the internal dataset.
% Fig~\ref{fig:gen_samples} provides a few query samples.

% \begin{table*}[ht!]
%     \centering
%     \begin{tabular}{c|c|c|c|c|c|c|c}
%         \toprule
%         \multirow{2}{*}{Retriever} & \multirow{2}{*}{MS MARCO v2.1} & \multicolumn{5}{|c|}{MTEB (VIDORE V2)} & \multirow{2}{*}{Internal Data} \\
%         & & ESG & Eco & ESGHL & BioMed & Avg & \\
%         \midrule
%         Semantic & 87.62 & 9.99 & 9.75 & 9.69 & 15.55 & 11.25 & 76.52 \\
%         Lexical & 80.22 & 6.76 & 10.47 & 3.90 & 8.66 & 7.45 & 72.55 \\
%         Multimodal & - & 45.23 & 51.38 & 57.48 & 51.04 & 51.28 & 73.84 \\
%         \bottomrule 
%     \end{tabular}
%     \caption{Information Retrieval Benchmarks}
%     \label{tab:placeholder}
% \end{table*}

\begin{table*}[ht!]
    \centering
    \begin{tabular}{ll|c|ccccc|c}
        \toprule
        \multirow{3}{*}{Retriever} & \multirow{3}{*}{Rewriter} & Text & \multicolumn{6}{c}{Unstructured} \\
         &  & \multirow{2}{*}{MARCO} & \multicolumn{5}{c|}{MTEB (Vidore V2)} & \multirow{2}{*}{Internal} \\
        & & & ESG & Eco & ESGHL & BioMed & Avg & \\
        \midrule
        \multirow{7}{*}{\textit{Semantic}} & Raw query & 87.62 & 9.99 & 9.75 & 9.69 & 15.55 & 11.25 & \underline{76.52} \\
        & Qwen3-4B & 71.08 & 23.09 & 35.94 & 31.45 & 35.36 & 31.22 & 28.08 \\
        & + RL-QR (Ours) & \underline{92.43} & 28.34 & 47.27 & 25.29 & 33.42 & 33.58 & 74.86 \\
        & Qwen3-8B & 84.21 & 27.70 & \underline{50.72} & 30.47 & 36.71 & 36.40 & 27.86 \\
        & + RL-QR (Ours) & 91.92 & \textbf{32.15} & \textbf{51.69} & \textbf{33.76} & \textbf{41.54} & \textbf{39.79} & 73.97 \\
        & Qwen3-14B & 70.73 & \underline{29.60} & 49.28 & 31.77 & 26.82 & 34.37 & 69.00 \\
        & + RL-QR (Ours) & \textbf{92.67} & 28.97 & 49.22 & \underline{33.69} & \underline{37.69} & \underline{37.39} & \textbf{80.61} \\
        \midrule
        \multirow{7}{*}{\textit{Lexical}} & Raw query & 80.22 & 6.76 & 10.47 & 3.90 & 8.66 & 7.45 & 72.55 \\
        & Qwen3-4B & 61.46 & 9.13 & 31.72 & 10.50 & 22.78 & 18.53 & 21.26 \\
        & + RL-QR (Ours) & \underline{84.84} & \underline{12.57} & 29.31 & \textbf{11.50} & \textbf{33.79} & \underline{21.79} & 76.45 \\
        & Qwen3-8B & 71.15 & 11.25 & 36.29 & 9.53 & 20.56 & 19.41 & 23.67 \\
        & + RL-QR (Ours) & 82.85 & \textbf{15.17} & \underline{38.93} & \underline{10.75} & \underline{26.94} & \textbf{22.95} & \underline{78.76} \\
        & Qwen-14B & 58.43 & 8.72 & \textbf{39.70} & 9.67 & 20.18 & 19.57 & 60.03 \\
        & + RL-QR (Ours) & \textbf{85.10} & 12.44 & 38.66 & 8.14 & 24.72 & 20.99 & \textbf{80.84} \\
        \midrule
        \multirow{7}{*}{\textit{Multimodal}} & Raw query & - & 45.23 & 51.38 & 57.48 & 51.04 & 51.28 & 73.84 \\
        & Qwen3-4B & - & 24.17 & 32.08 & 37.30 & 23.49 & 29.26 & 68.19 \\
        & + RL-QR (Ours) & - & 48.48 & 41.46 & 41.95 & 37.52 & 42.35 & 78.23 \\
        & Qwen3-8B & - & 48.30 & 53.69 & 55.10 & 51.14 & 52.06 & 70.80 \\
        & + RL-QR (Ours) & - & \underline{55.44} & \underline{57.82} & \textbf{63.46} & \underline{59.29} & \underline{59.00} & 79.83 \\
        & Qwen3-14B & - & 10.80 & 8.28 & 19.84 & 12.00 & 12.73 & 77.54 \\
        & + RL-QR (Ours) & - & \textbf{58.83} & \textbf{62.43} & \underline{62.47} & \textbf{60.02} & \textbf{60.94} & \textbf{81.08} \\
        \bottomrule 
    \end{tabular}
    \caption{Information retrieval benchmark results with NDCG@3 metrics. We report performance across varying model sizes (4B, 8B, 14B) to demonstrate the robustness of RL-QR. Note that RL-QR consistently improves performance over Raw Query even with smaller model backbones.}
    \label{tab:quant}
\end{table*}
\begin{table}[ht!]
    \centering
    \begin{tabular}{l|l|l}
        \toprule
        Retriever & Rewriter & Length \\ % & Gain \\
        \midrule
        \multirow{3}{*}{\textit{Semantic}} & Raw query & $36 \pm 14$ \\ % & 0\\
        & Qwen3-14B & $95 \pm 104 $ \\ % & -17 \\
        & +RL-QR & $38 \pm 15$ \\ % & +5 \\
        \midrule
        \multirow{3}{*}{\textit{Lexical}} & Raw query & $36 \pm 14$ \\% & 0 \\
        & Qwen3-14B & $95 \pm 104$ \\%& -22 \\
        & +RL-QR & $36 \pm 14$ \\%& +5 \\
         \bottomrule
    \end{tabular}
    \caption{Query length statistics on MS MARCO benchmark.} %  Gain refers to NDCG retrieval performance gain with respect to the raw query.
    \label{tab:stat_marco}
\end{table}
\begin{table}[ht!]
    \centering
    \begin{tabular}{l|l|l}
        \toprule
        Retriever & Rewriter & Length \\%& Gain \\
        \midrule
        \multirow{3}{*}{\textit{Semantic}} & Raw query & $84 \pm 23$ \\%& 0\\
        & Qwen3-14B & $157 \pm 199$ \\%& +23 \\
        & +RL-QR & $164 \pm 113$ \\%& +26 \\
        \midrule
        \multirow{3}{*}{\textit{Lexical}} & Raw query & $84 \pm 23$ \\%& 0 \\
        & Qwen3-14B & $190 \pm 258$ \\%& +12 \\
        & +RL-QR & $143 \pm 168$ \\%& +14 \\
        \midrule
        \multirow{3}{*}{\textit{Multimodal}} & Raw query & $84 \pm 23$ \\%& 0 \\
        & Qwen3-14B & $43 \pm 171$ \\%& -39 \\
        & +RL-QR & $116 \pm 48$ \\%& +10 \\
         \bottomrule
    \end{tabular}
    \caption{Query length statistics on MTEB Vidore V2 benchmark.}
    \label{tab:stat_mteb}
\end{table}
\begin{figure*}[ht!]
    \centering
    \small
    \begin{tabular}{p{2.5cm}|p{12.5cm}}
        \toprule
        \multicolumn{2}{c}{\large{\centering{MS MARCO v2.1 @ Semantic Retriever}}} \\
        Raw query & (0.47) cost of finishing basement which is half done' \\
        Qwen3-14B & (0.30) Average cost to finish a partially completed basement \\
        RLQR-14B & (0.70) cost to finish a basement that is halfway completed \\
        \hline
        Raw query & (0.47) how is chicken distributed to consumers \\
        Qwen3-14B & (0.00) Explain the process of chicken distribution from farms to consumers, including supply chain, transportation, packaging, and retail channels. \\
        RLQR-14B & (0.77) chicken distribution to consumers \\
        \midrule
        \multicolumn{2}{c}{\large{\centering{MS MARCO v2.1 @ Lexical Retriever}}} \\
        Raw query & (0.77) computer networking definition and examples \\
        Qwen3-14B & (0.77) Explain the concept of computer networking and provide real-world examples of its applications \\
        RLQR-14B & (1.0) computer network definition and examples \\
        \hline
        Raw query & (0.0) granby missouri high and low temp by month calendar \\
        Qwen3-14B & (0.0) Monthly average high and low temperatures in Granby, Missouri, with climate data organized by month for a yearly temperature calendar \\
        RLQR-14B & (0.47) granby missouri monthly high and low temperature calendar \\
        \midrule
        \multicolumn{2}{c}{\large{\centering{MTEB VIDORE V2 @ Semantic Retriever}}} \\
        Raw query & (0.0) What are the specific outcomes of using autologous chondrocyte implantation in canine studies? \\
        Qwen3-14B & (0.0) What are the clinical outcomes, efficacy, and long-term results of autologous chondrocyte implantation in canine cartilage repair studies, including veterinary research and experimental trials? \\
        RLQR-14B & (0.47) What are the clinical outcomes, efficacy, and complications associated with autologous chondrocyte implantation (ACI) in canine cartilage repair studies? \\
        \hline
        Raw query & (0.0) What are the differences in results obtained from two-photon microscopy and confocal microscopy? \\
        Qwen3-14B & (0.0) What is the difference between two-photon microscopy and confocal microscopy in terms of their results? \\
        RLQR-14B & (1.0) What are the key differences in resolution, penetration depth, phototoxicity, and image quality between two-photon microscopy and confocal microscopy results? \\
        \midrule
        \multicolumn{2}{c}{\large{\centering{MTEB VIDORE V2 @ Lexical Retriever}}} \\
        Raw query & (0.0) What role do biomaterials play in the development of joint replacement prostheses? \\
        Qwen3-14B & (0.0) What are the key contributions of biomaterials to the design, functionality, and longevity of joint replacement prostheses, including their biocompatibility and interaction with human tissues? \\
        RLQR-14B & (0.31) What is the role and function of biomaterials in the design and development of joint replacement prostheses, including their biocompatibility, durability, and contribution to tissue integration? \\
        \hline
        Raw query & (0.0) How do the roles of integrins in cell-matrix interactions compare to their roles in signal transmission within the cell? \\
        Qwen3-14B & (0.63) What are the differences in the roles of integrins in extracellular matrix (ECM) interactions versus their involvement in intracellular signaling pathways and mechanisms? \\
        RLQR-14B & (1.0) Compare the functions of integrins in cell-extracellular matrix interactions with their roles in intracellular signal transduction pathways. \\
        \midrule
        \multicolumn{2}{c}{\large{\centering{MTEB VIDORE V2 @ Multimodal Retriever}}} \\
        Raw query & (0.23) What role do scaffolds play in tissue engineering? \\
        Qwen3-14B & (0.30) What is the role of a scaffold in tissue engineering and why is it important? \\
        RLQR-14B & (0.47) What is the function and importance of scaffolds in tissue engineering, including their structural support, cell interaction, and role in tissue regeneration \\
        \hline
        Raw query & (0.0) What are the key factors influencing bioadhesion in biomaterials? \\
        Qwen3-14B & (0.5) What is bioadhesion in the context of biomaterials, and what factors determine its effectiveness? \\
        RLQR-14B & (0.63) What are the primary factors that affect bioadhesion in biomaterials, including molecular interactions, surface properties, and biological compatibility? \\
        \bottomrule
    \end{tabular}
    \caption{Qualitative results with the raw query, the baseline rewriting model, and the proposed method rewriter, RL-QR. Each sample is in form of `(NDCG@3 score) query'. }
    \label{fig:qual}
\end{figure*}

\subsection{Reinforcement Learning Query Rewriter}
We initialize the rewriter model with Qwen3 in \verb|no_think| mode.
We train them by the objective single epoch on eight 80GB H100 GPUs without any supervised finetuning.
Each training took 1 to 2 hours and the most bottlenecks are the retrieval overhead and the conversation between the search engine and the rewriter.
For the GRPO RL training, we adopt \verb|TRL| library of \verb|huggingface| with \verb|deepspeed| and its default training settings: learning rates, the number of rollouts and more. 

\section{Results}
\label{sec:results}

The experiments are designed to systematically compare the retrieval performance between (1) the raw query, (2) the vanilla base model Qwen3, and (3) the RL-QR trained models.
Our primary goal is to isolate the performance gains directly attributable to the proposed RL-QR training strategy.

\subsection{Overall Performance Comparison}
\label{subsec:overall}

As shown in Table~\ref{tab:quant}, the vanilla Qwen3 base model exhibits inconsistent performance.
It could achieve retrieval gains only on the MTEB-\textit{Semantic} and MTEB-\textit{Lexical} tasks, showing poorer performance than the raw queries in all other evaluated settings.

On the other hand, the proposed \textbf{RL-QR} method attains robust improvements across \textit{all} index-retriever cases.
RL-QR demonstrates substantial gains, obtaining improvements up to $3.9\times$ and from at least $5\%$.
This consistent superiority across diverse retrieval tasks strongly indicates that the performance improvements are a direct result of the proposed RL-QR approach, not merely the inherent strength of the baseline models.

\subsection{Domain-Specific Performance}
\label{subsec:domain_perf}

\textbf{Text-Modal Benchmark:} On the MS MARCO v2.1 experiment, RL-QR advances the retrieval performance by more than $5\%$ on both the \textit{Semantic} and \textit{Lexical} retrievers.
This is in stark contrast to the baseline rewriter, which fails significantly, degrading performance by up to $-27\%$ in the worst case and $-4\%$ in the best case.
For the semantic retriever (Qwen3-14B), RL-QR achieves an NDCG@3 of 92.67 compared to 87.62 for the raw query.

\textbf{Unstructured Multimodal Benchmarks:} RL-QR notably boosts the recall for the unstructured multimodal settings.
We observe gains of $3.9\times$ and $3.5\times$ on \textit{Lexical}-VidoreV2 and \textit{Semantic}-VidoreV2, respectively, with improvements ranging from $5\%$ to $20\%$ for the other combinations.
The baselines partially enhance performance on specific tasks but suffer dramatic degradation on others, such as a $-71\%$ drop on \textit{Lexical}-Internal.

\subsection{Qualitative Analysis of Rewriting Strategies}
\label{subsec:qualitative}

To understand the mechanism behind the performance gains, we analyzed the rewritten queries shown in Figure~\ref{fig:qual}.
We observe that RL-QR does not apply a uniform rewriting policy; rather, it adapts its strategy based on the domain and the nature of the retriever.

\textbf{Query Refinement vs. Expansion:} In the MS MARCO benchmark (Web Search), RL-QR primarily functions as a query \textit{refiner}.
As seen in the ``basement cost'' example, the model corrects grammatical structures and sharpens the intent ($0.70$ NDCG) without significantly increasing length.
Conversely, in the MTEB VIDORE V2 benchmarks (Technical/Medical), the model shifts to \textit{query expansion}.
For instance, when querying about ``autologous chondrocyte implantation,'' RL-QR explicitly injects related technical terms such as ``clinical outcomes,'' ``efficacy,'' and ``complications''.
This expansion, which achieved a $0.47$ NDCG score compared to $0.00$ for the baseline, bridges the vocabulary gap in specialized corpora.

\subsection{Behavioral Alignment for Retrieval}
\label{subsec:alignment}

A critical failure mode of the baseline Qwen3 model is its tendency to interpret search queries as generative instructions.
As shown in Figure~\ref{fig:qual}, for the query ``how is chicken distributed,'' the baseline generates a verbose command: ``Explain the process of chicken distribution...'' resulting in a score of $0.00$.
This suggests the pre-trained model is misaligned for retrieval tasks, prioritizing conversational fluency over keyword matching.

In contrast, RL-QR effectively ``unlearns'' this chat-oriented behavior.
It strips away conversational artifacts and focuses on keyword density and search intent.
For the same chicken distribution query, RL-QR outputs ``chicken distribution to consumers'' ($0.77$ NDCG), demonstrating that the RL optimization successfully realigned the model's output distribution from \textit{instruction following} to \textit{index-oriented retrieval}.

\subsection{Effect of Query Length and Index-Awareness}
\label{subsec:length_index}

RL-QR achieved retrieval enhancements irrespective of whether the query length was preserved or enlarged.
Cross-referencing the length statistics in Table~\ref{tab:stat_mteb} with the samples in Figure~\ref{fig:qual} reveals a dynamic, index-aware adaptability.

In precision-oriented environments like MS MARCO, the model retains brevity, with query lengths remaining close to the raw input (e.g., $38 \pm 15$ tokens).
For example, for the ``granby missouri'' query, it merely reorders keywords to match a likely document title format ($0.47$ NDCG), avoiding the excessive verbosity seen in the baseline ($0.00$ NDCG).

Conversely, in recall-oriented multimodal environments, RL-QR significantly expands the query length ($116 \pm 48$ tokens).
For the multimodal query on ``scaffolds in tissue engineering,'' RL-QR hallucinates visual and structural context (``structural support,'' ``cell interaction''), boosting the score from $0.23$ to $0.47$.
This behavior results from the training objective (Eq.~\ref{eq:grpo}), which does not explicitly penalize length, allowing the model to learn an implicit representation of the corpus distribution and adjust its verbosity accordingly.
\section{Conclusion}
In this work, we presented RL-QR, an annotation-free reinforcement learning framework for query rewriting that eliminates the need for expensive human-annotated training data in Retrieval-Augmented Generation systems. 
By synthesizing index-aligned queries and directly optimizing the rewriter with verifiable search rewards derived from NDCG, RL-QR achieves robust and substantial retrieval improvements across diverse retrievers and modalities—up to 3.9× on lexical and 3.5× on semantic retrievers for unstructured visual documents on the MTEB VIDORE V2 benchmark, alongside consistent gains of 5\%--10\% on MS MARCO v2.1 and internal industrial datasets.

The proposed approach is retriever- and index-agnostic, modular, and readily deployable in production environments, significantly reducing the maintenance overhead associated with domain-specific retriever tuning or re-indexing. 
Qualitative analysis further reveals that RL-QR adaptively learns index-aware rewriting strategies—ranging from concise refinement in text-heavy domains to targeted expansion in technical and multimodal corpora—effectively aligning user queries with the representation space of the underlying index.

RL-QR thus offers a scalable, cost-effective solution to one of the central bottlenecks in modern RAG systems. 
Future directions include extending the framework to multi-turn conversational retrieval, incorporating richer reward signals from downstream generation quality, and exploring its integration with emerging multimodal foundation models.

% \clearpage

\bibliography{latex/custom}

% \appendix

% \section{Example Appendix}
% \label{sec:appendix}

% This is an appendix.

\end{document}